\pdfoutput=1
\documentclass[sigconf]{acmart}

\renewcommand\footnotetextcopyrightpermission[1]{} 

\usepackage{tablefootnote}
\newcommand*{\redbf}[1]{\textbf{\color{red}{#1}}}

\AtBeginDocument{%
  \providecommand\BibTeX{{%
    \normalfont B\kern-0.5em{\scshape i\kern-0.25em b}\kern-0.8em\TeX}}}

\setcopyright{acmcopyright}
\copyrightyear{2022}
\acmYear{2022}

\acmConference[WebSci '22]{14th ACM Web Science Conference 2022}{June 26--29, 2022}{Barcelona, Spain}
\acmBooktitle{14th ACM Web Science Conference 2022 (WebSci '22), June 26--29, 2022, Barcelona, Spain}
\acmPrice{15.00}

\acmDOI{10.1145/3501247.3531574}
\acmISBN{978-1-4503-9191-7/22/06}

\begin{document}

\title{Should we tweet this? Generative response modeling for predicting reception of public health messaging on Twitter}

\author{Abraham Sanders}
\orcid{0000-0002-5231-2239}
\affiliation{
  \institution{Rensselaer Polytechnic Institute}
  \city{Troy}
  \state{New York}
  \country{USA}
}
\email{sandea5@rpi.edu}

\author{Debjani Ray-Majumder}
\orcid{0000-0001-6433-9490}
\affiliation{
  \institution{Rensselaer Polytechnic Institute}
  \city{Troy}
  \state{New York}
  \country{USA}
}
\email{raymad@rpi.edu}

\author{John S. Erickson}
\orcid{0000-0003-3078-4566}
\affiliation{
  \institution{Rensselaer Polytechnic Institute}
  \city{Troy}
  \state{New York}
  \country{USA}
}
\email{erickj4@rpi.edu}

\author{Kristin P. Bennett}
\orcid{0000-0002-8782-105X}
\affiliation{
  \institution{Rensselaer Polytechnic Institute}
  \city{Troy}
  \state{New York}
  \country{USA}
}
\email{bennek@rpi.edu}

\renewcommand{\shortauthors}{Sanders, et al.}

\begin{abstract}
  The way people respond to messaging from public health organizations on social media can provide insight into public perceptions on critical health issues, especially during a global crisis such as COVID-19. It could be valuable for high-impact organizations such as the US Centers for Disease Control and Prevention (CDC) or the World Health Organization (WHO) to understand how these perceptions impact reception of messaging on health policy recommendations. We collect two datasets of public health messages and their responses from Twitter relating to COVID-19 and Vaccines, and introduce a predictive method which can be used to explore the potential reception of such messages. Specifically, we harness a generative model (GPT-2) to directly predict probable future responses and demonstrate how it can be used to optimize expected reception of important health guidance. Finally, we introduce a novel evaluation scheme with extensive statistical testing which allows us to conclude that our models capture the semantics and sentiment found in actual public health responses.
\end{abstract}

\ccsdesc[500]{Computing methodologies~Discourse, dialogue and pragmatics}
\ccsdesc[500]{Information systems~Sentiment analysis}

\keywords{tweet response generation, sentiment analysis, public health, COVID-19, vaccines}

\maketitle
\pagestyle{plain} 

\section{Introduction}\label{section_introduction}
During the COVID-19 pandemic, Twitter and other social media messaging by public health organizations played a significant role in their strategies to enact proposed mitigations to potential risks with varying effectiveness
\cite{WANG2021106568}. As such,  recent works have focused on topical, semantic, and sentiment analysis of COVID-19 and vaccine related Twitter discourse, many leveraging natural language processing (NLP) technologies. For example, Sanders et al. \cite{sanders2021unmasking} clustered tweets relating to mask-wearing in the early days of the COVID-19 pandemic to discover prevalent themes, perceptions, and sentiments. Cotfas et al. \cite{9354776} applied machine learning for vaccine stance detection using tweets collected in the month following the  announcement of a COVID-19 vaccine.  Our study follows similar motivation - to investigate the way the general population reacts to messaging from major public health agencies (e.g., US CDC, European CDC, and WHO) on a variety of topics including COVID-19 and vaccines. Unlike previous work in this area, we investigate the feasibility and utility of using state-of-the-art \textit{text generation models} to directly simulate typical response distributions to novel public health messages on Twitter. These simulations, combined with sentiment analysis, can be used to help public health organizations understand the specific opinions and concerns of their audience in order to develop more effective health messaging strategies. 

In this study, we collect two datasets of public health tweets: (1) COVID-19 related public health messages from March 1st, 2020 to September 30th, 2020, and (2) vaccine-related public health messages from October 1st, 2021 to January 31st, 2022. These datasets include the original messages and samples of their responses, both in the form of direct replies and quote-tweets (retweets with comments). Using each dataset, we fine-tune a GPT-2 \cite{radford2019language} language model to predict responses to the public health tweets and evaluate its effectiveness in terms of semantic and sentiment similarity with known responses. To evaluate the models, we establish ``ground-truth'' baselines for semantics and sentiment on each dataset by comparing two distinct samples of known responses to each message. We also establish ``random-chance'' baselines by likewise comparing each sample of known responses with a sample of random responses for any message in each dataset. We then use our models to generate responses to each test message compare them with the known response samples.
Through rigorous statistical testing we find that our models are able to generate responses consistent with known samples in terms of semantics and sentiment. Thus, insights on perceptions toward particular public health issues can be gained from analyzing the generated response distributions. We envision our methods being able to aid public health decision makers and social media content managers proactively model how the public will react to future messages, increasing the likelihood that their tweets are well received and have the intended impact.

The remainder of this paper is organized as follows:
(1) we present two datasets of Twitter public health messages and their responses, one related to COVID-19 and one related to Vaccines; (2) we fine-tune GPT-2 to generate responses on each of these datasets, and construct upper (ground-truth) and lower (random-chance) bound baselines against which to evaluate it; (3) we visually demonstrate the capabilities of our models using test set examples and walk through our envisioned public health use case; (4) we perform extensive statistical testing to compare our models against the baselines, finding that GPT-2 can effectively capture semantics and sentiment in typical response distributions to messages in our test sets; and (5) we conclude with a discussion of limitations and future directions of our work including a review of related works from the natural language generation (NLG) literature. 
We have released our data and code on GitHub, \footnote{Available at \url{https://github.com/TheRensselaerIDEA/generative-response-modeling}}
and, in compliance with the Twitter content redistribution policy,\footnote{See \url{https://developer.twitter.com/en/developer-terms/agreement-and-policy}} we only publish the tweet IDs corresponding to the actual tweets used in this work.

\section{Data Collection}\label{section_data_collection}

As in  \cite{sanders2021unmasking}, we used the Twitter streaming API to collect a random sample of tweets during the collection periods for each respective dataset (COVID-19 \& Vaccine public health messages). We collected these datasets by filtering the streaming API using COVID-19 and Vaccine related keywords, respectively. Since we aim to study the response distributions to public health tweets, we focus only on those tweets which have responses either in quote-tweet or direct reply form. Collection of these tweets and their responses was done via Tweepy, a python library for accessing the Twitter API, and they were stored in Elasticsearch\footnote{See \url{https://www.elastic.co/elasticsearch}} for efficient search and retrieval. Each dataset was then filtered by screen name to include only tweets from public health organizations and their responses. The organizations selected and their respective accounts are shown in Table \ref{ph_account_screen_names}.

\begin{table}[t]
    \centering
    \scriptsize
    \caption{\small Public health account screen names}
    \label{ph_account_screen_names}
    \begin{tabular}{l}
        \toprule
        \footnotesize
        \textbf{European CDC Accounts} \tablefootnote{See \url{https://www.ecdc.europa.eu/en/about-us/press-and-media/ecdc-social-media}} \\ 
        \midrule
        ESCAIDE ECDCPHT ecdc\_tb ECDC\_VPD ECDC\_HIVAIDS ecdc\_flu ECDC\_Outbreaks ecdc\_eu \\
		\toprule
        \footnotesize
        \textbf{U.S. CDC Accounts} \tablefootnote{See \url{https://www.cdc.gov/socialmedia/tools/Twitter.html}} \\ 
        \midrule
		cdcgov cdcdirector CDC\_eHealth CDCespanol BRFSS CDCasthma CDC\_DASH CDCDiabetes\\
		cdc\_drh CDCEnvironment CDC\_Cancer CDC\_EIDjournal CDC\_EPHTracking CDC\_Genomics\\
		CDC\_HIVAIDS CDCMicrobeNet CDC\_NCBDDD CDC\_NCEZID CDC\_TB CDC\_AMD\\
		CDCChronic CDCEmergency CDCFlu CDCGlobal CDCGreenHealthy CDCHaiti \\
		CDCHeart\_Stroke US\_CDCIndia CDChep CDCInjury CDCKenya CDCMakeHealthEZ \\
		CDCMMWR CDCNPIN CDCObesity cdcpcd CDCRwanda CDCsouthafrica CDCSTD \\ 
		CDCTobaccofree CDCTravel CPSTF DrDeanCDC DrKhabbazCDC DrMartinCDC \\ 
		DrMerminCDC DrNancyM\_CDC DrReddCDC MillionHeartsUS NCHStats niosh NIOSHMining \\
		NIOSH\_MVSafety NIOSH\_NPPTL NIOSH\_TWH nioshbreathe NIOSHConstruct NIOSHespanol \\
		NIOSHFACE nioshfishing nioshnoise NIOSHoilandgas WTCHealthPrgm\\
		\toprule
		\footnotesize
        \textbf{Other Public Health Accounts} \\ 
        \midrule
		WHO   InjectionSafety  \\
        \bottomrule
    \end{tabular}
    \vskip -0.1in
\end{table}

\subsection{COVID-19 Public Health Messaging} \label{section_covid_19_public_health_messaging}
\begin{figure}[t]
    \begin{center}
        \hspace*{-0.4cm}
        \includegraphics[width=1.07\columnwidth]{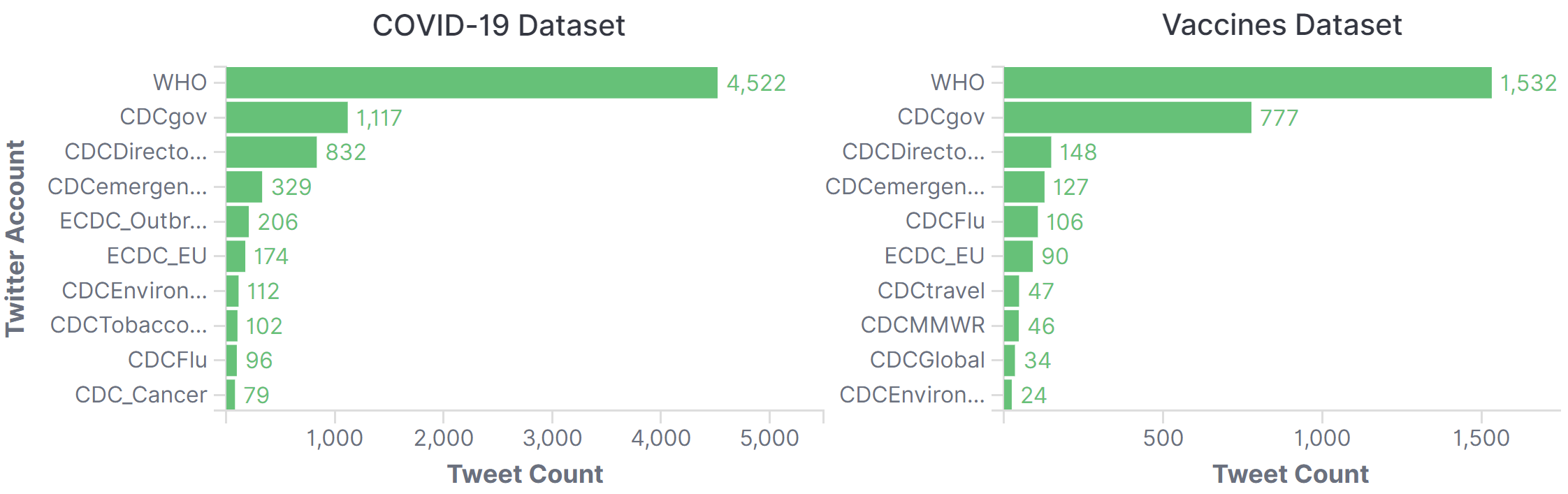}
        \caption{\small Top ten public health accounts by number of tweets for the COVID-19 and Vaccines datasets.}
        \label{public-health-accounts}
    \end{center}
    \vskip -0.05in
\end{figure}

Our dataset of COVID-19 related public health messages and their responses contains 8,475 original messages authored by these accounts and 70,331 responses to these messages. The original messages were authored between March 1st, 2020 and September 30th, 2020. The majority of the collected tweets originate from the WHO account, followed by CDCgov, as seen in Figure \ref{public-health-accounts}. This data was collected as follows: (1) We collected 295,468,580 original tweets from the Twitter Streaming API over the collection period using the same set of COVID-19 related filter keyphrases as used in \cite{sanders2021unmasking}; (2) These tweets were filtered to keep only those that were in response to (either via quote or direct reply) a message from one of the public health accounts in Table \ref{ph_account_screen_names}; (3) As the streaming API returned quoted tweets but not (direct) replied-to tweets, these were separately requested using the Twitter Status API.

\subsection{Vaccine Public Health Messaging} \label{section_vaccine_public_health_messaging}
Our dataset of Vaccine related public health messages and their responses contains 3,060 original messages authored by the accounts in Table \ref{ph_account_screen_names} and 61,009 responses to these messages. The original messages were authored between October 1st, 2021 and January 31st, 2022. The  majority of the collected tweets originate from the WHO account, followed by CDCgov, as is the case in the COVID-19 dataset (see Figure \ref{public-health-accounts}). This dataset was collected by the same procedure outlined for the COVID-19 dataset in Section \ref{section_covid_19_public_health_messaging}, with the only difference being the filter keyphrases. Here, all filter keyphrases were vaccine related, selected by doing a term-frequency analysis on a random sample of approximately 10,000 tweets collected using the keyphrase ``vaccine'' (see our code release for complete listing). Using these keyphrases we collected 52,282,174 original tweets before filtering for responses to the public health accounts.

\section{Experimental Setup}\label{section_experimental_setup}
As discussed in Section \ref{section_introduction}, we train GPT-2 on the task of tweet response generation. This task is notably different from other text generation tasks in that it suffers from an extreme form of the one-to-many problem seen in dialogue response generation, where an utterance can have many equally valid responses \cite{galley-etal-2015-deltableu, gupta-etal-2019-investigating, 8682634}.  Specifically, each public health message in our datasets has multiple responses, and we train GPT-2 to model the \textit{distribution} of typical responses for each message. This means that the same message from the same author is repeated many times in the training set, each instance with a different target response. Once trained in this manner, temperature sampling can be used to generate a range of likely responses to an input author and message.

As previously mentioned, we evaluate this method by comparing model-generated responses to known responses. Specifically, given a known sample of responses to a particular message and author, we need to determine how well a model-generated sample of responses captures the semantics (e.g., meaning, topic, intent) and the sentiment polarity (e.g., positive, negative, neutral) of the known responses. This is akin to measuring retrieval \textit{recall} - how well the model-generated response distribution ``covers'' that of the ground-truth. To measure sentiment we use a publicly available RoBERTa \cite{DBLP:journals/corr/abs-1907-11692} model\footnote{Obtained from \url{https://huggingface.co/cardiffnlp/twitter-roberta-base-sentiment}} fine-tuned on the sentiment classification task of the TweetEval benchmark \cite{barbieri-etal-2020-tweeteval}. We score the sentiment of each message and response in our datasets in the range $[-1, 1]$ by multiplying the sentiment class probabilities predicted by RoBERTa for negative, neutral and positive by $\{-1, 0, 1\}$ respectively and summing the result. To measure semantic similarity we compute sentence embeddings for each message and response in our datasets, and measure cosine similarity between embeddings. To compute the embeddings we use a publicly available MiniLM \cite{wang2020minilm} model\footnote{Obtained from \url{https://huggingface.co/sentence-transformers/all-MiniLM-L12-v2}} fine-tuned for semantic textual similarity using a contrastive objective on over one billion training pairs from 32 distinct datasets.

We now provide details of our experimental set up.

\subsection{Train / Test Split}\label{section_train_test_split}

\begin{table}[t]
    \centering
    \footnotesize
    \caption{\small Training \& Test set statistics}
    \label{training_test_set_statistics}
    \begin{tabular}{llll}
        \toprule
        \multicolumn{4}{l}{\textbf{COVID-19 Public Health dataset}}\\
        \midrule
        \textbf{Set} & \textbf{\# of messages.} & \textbf{\# of responses} & \textbf{Avg. \# resp. per msg.} \\
        \midrule
        Train & 7,860 & 39,020 & $5\pm 8^*$ \\
        Test & 155 & 9,300 & $60^{**}$\\
        \toprule
        \multicolumn{4}{l}{\textbf{Vaccine Public Health dataset}}\\
        \midrule
        \textbf{Set} & \textbf{\# of messages.} & \textbf{\# of responses} & \textbf{Avg. \# resp. per msg.} \\
        \midrule
        Train & 2790 & 27,878 & $10\pm 16^*$ \\
        Test & 140 & 8,400 & $60^{**}$ \\
        \bottomrule
        \multicolumn{4}{l}{\scriptsize *: Mean $\pm$ standard deviation, both rounded to the nearest integer.}\\
        \multicolumn{4}{l}{\scriptsize **: Exactly 60 responses were sampled for each message in the test set.}
    \end{tabular}
    \vskip -0.1in
\end{table}

For each dataset, we  set aside a test set of public health messages including all messages with at least 60 responses. For all experiments we choose a sample size of 30 responses, ensuring that we can randomly select
two distinct samples for the ground-truth baseline. We  clean the message text by removing hyperlinks and emojis, and remove all messages that are duplicated by the same author. This last step is taken since responses to duplicated messages often depend on external context beyond the message itself such as a hyperlink or embedded entity which may vary between the duplicates. As such,  a model trained on message text alone is unlikely to accurately predict responses to such messages.

After setting aside the test set, the remainder of the message-author-response triples in each dataset are used for fine-tuning GPT-2. As done for the test set, we clean the message and response text by removing hyperlinks and emojis, and remove duplicated messages from the same authors. Unlike the test set, we allow one instance of each duplicated message (along with its responses) to remain in the training set. As a final step, we remove any remaining message from the training set that is identical in content to a message in the test set. Statistics for the training and test sets for the COVID-19 and Vaccine datasets are provided in Table \ref{training_test_set_statistics}.

\subsection{Response Generation Model}\label{section_response_generation_model}
We then fine-tune the 762 million parameter GPT-2 model\footnote{Obtained from \url{https://huggingface.co/gpt2-large}} on the response generation task. Each training example consists of a public health message, the author account's screen name, and one response, delimited by three special tokens we add to the model's vocabulary: (1) \textbf{<|message|>} to indicate the following text is the public health message; (2) \textbf{<|author|>} to indicate the following text is the screen name of the message author; and (3) \textbf{<|response|>} to indicate the following text is a response to the message. At inference time, this enables generated response samples to be conditioned on the message text and author by prompting GPT-2 with the message and author followed by a \textbf{<|response|>} token as seen in Table \ref{gpt2_input_examples}.

\begin{table}[t]
    \centering
    \scriptsize
    \caption{\small GPT-2 training \& inference examples}
    \label{gpt2_input_examples}
    \begin{tabular}{l}
        \toprule
        \footnotesize
        \textbf{Training Examples:}\\
        \midrule
        \redbf{<|message|>}Is your child worried about \#COVID19? Learn the facts so you can answer your \\
        children's questions. Make sure to explain the simple things they can do, like washing their \\
        hands often. Learn more here:\redbf{<|author|>}CDCgov\redbf{<|response|>}Great resource for parents and \\
        teachers. \#COVID19 \#ProtectKidsHealth\redbf{<|endoftext|>} \\
        \midrule
        \redbf{<|message|>}How children can \#WearAMask properly. Read more about children \& masks in \\
        relation to \#COVID19\redbf{<|author|>}WHO\redbf{<|response|>}How long back into school until kids are \\
        trading their avengers masks like pogs?\redbf{<|endoftext|>} \\
        \midrule
        \redbf{<|message|>}How children can \#WearAMask properly. Read more about children \& masks in \\
        relation to \#COVID19\redbf{<|author|>}WHO\redbf{<|response|>}Children do not need to wear masks! \\ 
        Get lost!!\redbf{<|endoftext|>} \\
        \toprule
        \footnotesize
        \textbf{Inference Example (not actual model output):}\\
        \midrule
        \textbf{Prompt:} \redbf{<|message|>}How will people respond to THIS?\redbf{<|author|>}CDCdirector\redbf{<|response|>}\\
        \textbf{Output sample 1:}\ \ Like this! \redbf{<|endoftext|>} \\
        \textbf{Output sample 2:}\ \ And this! \redbf{<|endoftext|>} \\
        \textbf{\dots}\\
        \textbf{Output sample N:}\ \#DefinitelyThis. \redbf{<|endoftext|>} \\
        \bottomrule
    \end{tabular}
    \vskip -0.05in
\end{table}

Before fine-tuning, 10\% of the training set is held out as a validation set. Fine-tuning is then done with the AdamW optimizer \citep{loshchilov2018decoupled} with an initial learning rate of $3 \times 10^{-5}$ for a maximum of 15 epochs. Validation and checkpointing is done 4 times per epoch, and training is terminated early if three epochs elapse with no improvement in validation loss. Once training completes, the checkpoint corresponding to the lowest validation perplexity is selected as the final model. We train separate GPT-2 models on the COVID-19 and Vaccine datasets and report training statistics for both in Table \ref{training_statistics}.

After training, each fine-tuned model is used to generate 30 responses to each message in its respective test set. All generation is done with beam sampling using num\_beams=3, top\_k=50, top\_p=0.95, and temperature=1.5.

\begin{table}[t]
    \centering
    \footnotesize
    \caption{\small Model training statistics}
    \label{training_statistics}
    \begin{tabular}{lllll}
        \toprule
        \textbf{Dataset} & \textbf{\# of train ex.} & \textbf{\# of val ex.} & \textbf{\# of epochs} & \textbf{Final val PPL.} \\
        \midrule
        COVID-19 & 35,118 & 3,902 & 2 & 3.36 \\
        Vaccines & 25,090 & 2,788 & 2 & 2.82 \\
        \bottomrule
        \multicolumn{5}{l}{\scriptsize ex = examples; val = validation; PPL = perplexity}
    \end{tabular}
    \vskip -0.05in
\end{table}

\begin{figure*}[t]
\begin{center}
\includegraphics[width=\textwidth]{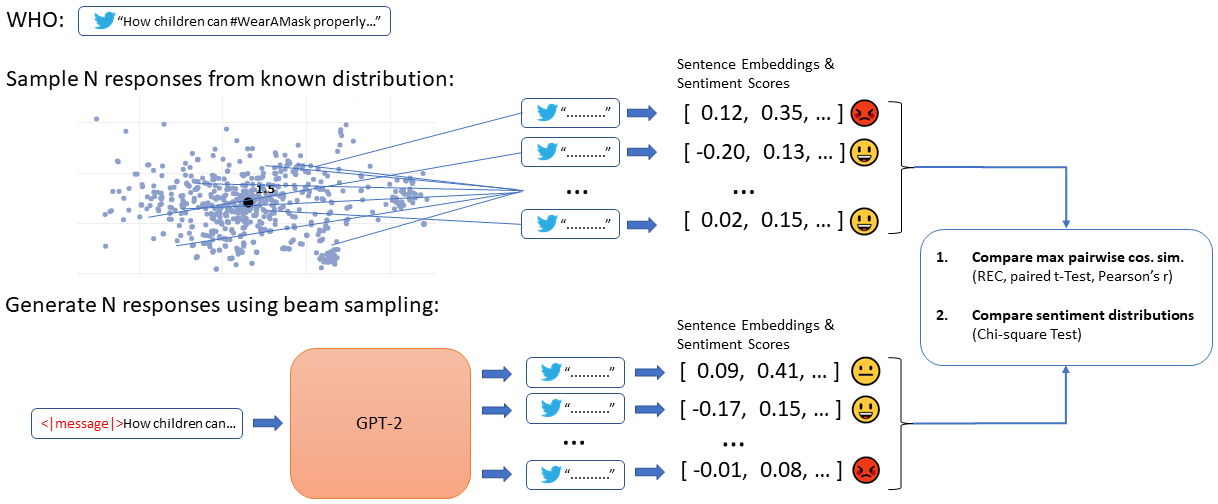}
\caption{\small An illustration of our model evaluation scheme for a single test message. For each test message, N=30 generated responses are compared with the primary ground-truth sample of N=30 known responses. The sentence embeddings and sentiment scores for each response are then used to perform statistical testing for semantic and sentiment similarity. The ground-truth and random-chance baselines are computed the same way, replacing the generated responses with the reference ground-truth and random response samples respectively.}
\label{experiment_setup}
\end{center}
\end{figure*}

\subsection{Evaluation \&  Baselines}\label{section_evaluation_baselines}
 
Finally, we use the test set of each dataset to establish the ground-truth and random-chance baselines which function as expected upper and lower bounds, respectively, for semantic and sentiment similarity measurements. For each message in the test set, we sample: (1) 60 known responses, and (2) 30 responses to random messages in the dataset. The 60 known responses are split into two distinct ``ground-truth'' sets - a \textbf{Primary} set and a \textbf{Reference} set used for establishing a baseline. Thus, for each test message we compare the 30 primary ground-truth responses with:
\begin{enumerate}
    \item the 30 reference responses (\textbf{ground-truth baseline}).
    \item the 30 model-generated responses (\textbf{model evaluation}).
    \item the 30 random responses (\textbf{random-chance baseline}).
\end{enumerate}
Figure \ref{experiment_setup} illustrates this evaluation scheme. As shown, we use several statistical tests to measure semantic and sentiment similarity for the baselines and for model evaluation. Details of these tests and their results are provided in Section \ref{section_quantitative_results}.

\section{Qualitative Results}\label{section_qualitative_results}

\subsection{Model Output Examples}\label{section_model_output_examples}

\begin{figure*}[t]
\begin{center}
\scriptsize
\begin{tabular}{l}
    \toprule
    \textbf{Message (CDCDirector; COVID-19 dataset):} \\
    Together, we can open schools safely. This fall, school may look different in some communities b/c of \#COVID19 as face masks, distancing, \& hand hygiene meet bookbags \& homework. \\
    \hspace{0.1in} New CDC tools will help us make decisions for the upcoming school year.\\
    \addlinespace
    \textbf{Primary ground-truth responses:} \\
    @CDCDirector @CDCgov \textcolor{purple}{Shame on you} and the @CDC. Trump canceled the @GOP convention, but it's just fine sending our kids back to school. The curve has NOT been flattened \\
    \hspace{0.1in} and we are trending in the wrong direction, but send kids back to school? \textcolor{blue}{What has happened to the once great CDC?} \textcolor{purple}{Shameful}.\\
    @CDCDirector @CDCgov Sad to see the once esteemed CDC \textcolor{orange}{become the patsy of an administration} that doesn't respect or understand data. Luckily, the American public does seem \\
    \hspace{0.1in} to understand death tolls. \\
    @CDCDirector @CDCgov \textcolor{red}{CDC is misleading America} parents and \textcolor{brown}{putting children's in great harm with its new guidelines for schools} during this pandemic. There's not enough data \\
    \hspace{0.1in} to support younger children's are less likely to contract and spread this virus. YOU ARE PLAYING GOD.\\
    @CDCDirector @CDCgov We the people are wondering if the \textcolor{red}{cdc is taking bribes to cook the books}....\\
    @CDCDirector @CDCgov But the Republican Convention is not safe? The CDC is supposed to protect the American people \textcolor{orange}{and not do one person's bidding.} \\
    \hspace{0.1in} Remember your Hippocratic oath do no harm\\
    \addlinespace
    \textbf{Generated responses:}\\
    @CDCDirector @CDCgov The \textcolor{red}{CDC has lost its credibility.} You should be removed immediately. You are not leading. \textcolor{orange}{You are a Trump puppet.} Resign.\\
    @CDCDirector \textcolor{brown}{How do you expect schools to be safe when you keep changing the testing recommendations?} You are not a doctor, you are a politician. You have \\
    \hspace{0.1in} \textcolor{red}{destroyed the credibility of the CDC}. Resign!\\
    @CDCDirector You should be removed from office immediately. \textcolor{purple}{You are a disgrace} \textcolor{blue}{to the once respected CDC.}\\
    @CDCDirector @CDCgov \textcolor{brown}{Why are you putting children at risk?} You say it will be safe, but you also say it will be deadly for children to be in schools. The CDC needs to be \\
    \hspace{0.1in} \textcolor{red}{dismantled and put in place someone who can be trusted}. \textcolor{purple}{You're a disgrace}.\\
    @CDCDirector @CDCgov Why is your department still pushing for more testing? \textcolor{purple}{You are a disgrace} \textcolor{blue}{to the once respected CDC} by not standing up for science and the American people. \\
    \hspace{0.1in} You are complicit in thousands of deaths! \textcolor{orange}{Stand up to Trump} and do your job!\\
    \midrule
    \textbf{Message (WHO; COVID-19 dataset):}\\
    ``\#COVID19 has taken so much from us, but it has also given us a unique opportunity: -To put aside our differences -To break down barriers -To see and seek the best in each other -To lift \\
    \hspace{0.1in} our voices for \#HealthForAll''-@DrTedros during the One world, \#TogetherAtHome\\
    \addlinespace
    \textbf{Primary ground-truth responses:} \\
    @WHO @DrTedros "\#COVID19 has taken so much from us, but it has also given us a unique opportunity: -To reveal the \textcolor{red}{corruption and incompetence of @WHO} -To have a glimpse to \\
    \hspace{0.1in} \textcolor{blue}{the dictatorship of the CCP} \\
    @WHO @DrTedros \textcolor{blue}{Why not let China fund your escapades?} You can't have solidarity with people or \textcolor{red}{an organization you don't trust}. \textcolor{blue}{\#ChinaLiedPeopleDied} \textcolor{red}{\#WHOLiedPeopleDied} \\
    \hspace{0.1in} \textcolor{red}{\#DefundWHO} \textcolor{blue}{\#CCPVirus \#XiJinping} \#winnietheflu\\
    @WHO @DrTedros Keep the meaningless platitudes coming, I'm sure that's the key to stopping this pandemic you opened the door to through your \textcolor{blue}{pandering to China}. Bang up job.\\
    @WHO @DrTedros Let's be inclusive and include ! Amazing that the country EXCLUDED from the @WHO handed the \textcolor{blue}{Wuhan Virus} the best.\\
    @WHO @DrTedros -To \textcolor{blue}{shill for china}\\
    \addlinespace
    \textbf{Generated responses:}\\
    @WHO @DrTedros How dare you chime in on a global pandemic with the help of China in the middle of a global economic crisis. The \textcolor{blue}{Chinese Communist Party should be held} \\
    \hspace{0.1in} \textcolor{blue}{fully accountable for this global tragedy}.\\
    @WHO @DrTedros \textcolor{blue}{\#ChinaLiedPeopleDied China should be brought up on charges} for this global pandemic. We need justice.\\
    @WHO @DrTedros \textcolor{red}{WHO} and \textcolor{blue}{CHINA LIED and PEOPLE DIED!} INVESTIGATE\\
    @WHO @DrTedros \textcolor{blue}{The Chinese are not your friend}. You should be kicked out of every country of the world. \textcolor{red}{You are the cause of this global crisis}.\\
    @WHO @DrTedros @DrTedros, \textcolor{red}{you must resign! You are a disgrace and a danger to the world!} The whole world is suffering from your failures! China is suffering from you!\\
    \midrule
    \textbf{Message (CDCgov; Vaccines dataset):}\\
    URGENT: If you are pregnant, recently pregnant, or plan to get pregnant, get vaccinated against \#COVID19 as soon as possible. Getting vaccinated can protect you from severe illness from \\ 
    \hspace{0.1in} COVID-19, and a healthy parent is important for a healthy baby. More:\\
    \addlinespace
    \textbf{Primary ground-truth responses:} \\
    @CDCgov My pregnant WIFE, will not be getting vaccinated. \textcolor{red}{Not enough studies on how it will affect our soon to be son!}\\
    It is alarming to see that only 31\% of pregnant people are vaccinated, but it is \textcolor{blue}{encouraging to see the CDC speak out about the importance of pregnant people receiving} \\ 
    \hspace{0.1in} \textcolor{blue}{their Covid-19 vaccine}. \#GburgGHFall21\\
    @CDCgov Do you not even care about the pregnant men that can die? What's the statistics on that? Why aren't you encouraging them to get the vaccine???\\
    Today: @CDC \textcolor{blue}{recommends \#COVID19Vaccine if you're pregnant}, recently pregnant, or plan to get pregnant. \#ArmyMedicine \textcolor{blue}{\#InThisTogether}\\
    @CDCgov @CDPHE \textcolor{brown}{How did the FDA categorize the Covid vaccines?} Category C? Has to be a C. Let's see those studies - oh wait you are doing them NOW! Criminal\\
    \addlinespace
    \textbf{Generated responses:}\\
    @CDCgov So why are pregnant women and people who have already had covid vaccinated against covid sick and dying?\\
    @CDCgov \textcolor{red}{What about the babies that are being born after these vaccines?}\\
    @CDCgov What about the pregnant women who get vaccinated and then get covid and die? The babies? Who's going to pay for this?\\
    The @CDCgov is telling pregnant women to get vaccinated, \textcolor{brown}{but the @US\_FDA has not authorized} or recommended it for pregnant women. When will the @CDCgov and @CDCDirector \\
    \hspace{0.1in} do their jobs and approve COVAXIN?\\
    \#Pregnant people: \textcolor{blue}{Get vaccinated against \#COVID19 if you're pregnant. \#Gratitude} \#WeArePalmGarden\\
    \midrule
    \textbf{Message (CDCDirector; Vaccines dataset):}\\
    Today, I endorsed ACIP's recommendation that children 5-11 yrs old should be vaccinated against \#COVID19 w/ Pfizer-BioNTech pediatric vaccine. This expands vaccine recommendations to \\
    \hspace{0.1in} over 28M kids in US \& now allows providers to begin vaccinating them.\\
    \addlinespace
    \textbf{Primary ground-truth responses:} \\
    @CDCDirector Do not vaccinate children with this cv vaccine. They have natural immunity and the vax doesn't work. HCG and Ivermetin does. Leave the Children alone you monsters\\
    It has happened. \textcolor{red}{I am grateful for your efforts in keeping out children safe!} \#cdc \#Covid19vaxxed \#COVIDVaccination \#safe\\
    All 5-11 year old - let go, line up and get vaccinated against \#Covid19. Help make this world \#Covid19 free!\\
    Children 5-11 years old are now authorized to receive the Pfizer COVID vaccine. \textcolor{red}{A huge step forward in our continued to efforts to combat COVID and keep our kids safe.}\\
    \addlinespace
    \textbf{Generated responses:}\\
    @CDCDirector @CDCgov What about the kids who are not eligible for the vaccine? When are you going to talk about that? What are you waiting for?\\
    @CDCDirector @ACIP When can kids under 5 get the vaccine? \textcolor{red}{It is the most important vaccine for them.}\\
    \textcolor{red}{Thank you @CDCgov for recommending this vaccine for children!} We now need to support the @US\_FDA and @NIH to approve this vaccine for 5-11 year olds. We can't simply hope for the \\
    \hspace{0.1in} pandemic to end on its own.\\
    @CDCDirector Please \textcolor{red}{Protect: Yourself Your Family Other People Get Fully \#Vaccinated} Against: \#COVID19 \#Influenza / Flu Measles, Mumps, Polio, HPV, etc Wear a Mask Public Places Public\\ 
    \hspace{0.1in} Transportation Wash Your Hands @CDCgov\\
    \bottomrule
\end{tabular}
\caption{\small Example generated responses to test set messages on both datasets. Elements of each response that share similar themes are highlighted with the same color (e.g., \textcolor{red}{red} text in the first example indicates mistrust and \textcolor{red}{red} text in the last example indicates gratefulness). We observe that the generated responses capture many of the same opinions and concerns as the primary (known) responses. }
\label{example_generated_responses}
\end{center}
\end{figure*}

In Figure \ref{example_generated_responses} we show primary ground-truth and model-generated responses for two messages from each test set (COVID-19 \& Vaccines). For each message, we show the top five ground-truth responses ranked in descending order of mean cosine similarity (defined in Section \ref{section_quantitative_results}) with the model-generated responses, and likewise we show the top five model-generated responses ranked in descending order of mean cosine similarity with the ground-truth responses. This filtering and ordering is done for the sake of brevity as it is not practical to fit all $60 \times 4$ responses in this document. We observe that the generated responses capture many of the same opinions and concerns present in the known responses. We summarize some of the key similarities evident in the examples: 

The first example shows a test message from  the COVID-19 dataset where CDCDirector recommends that schools can re-open safely. The known and generated responses both exhibit themes of mistrust toward the CDC (shown in \textcolor{red}{red}), allegations of bowing to the Trump administration (shown in \textcolor{orange}{orange}), implication of shame and disgrace toward the CDC (shown in \textcolor{purple}{purple}), concern for the well-being of school children (shown in \textcolor{brown}{brown}), and references to loss of former respect (shown in \textcolor{blue}{blue}). The second example shows a test message from the COVID-19 dataset where WHO calls for unity in the face of the pandemic. The known and generated responses both exhibit themes of mistrust toward the WHO (shown in \textcolor{red}{red}) and allegations of conspiracy with China (shown in \textcolor{blue}{blue}). The third example shows a test message from the Vaccines dataset where CDCgov urges pregnant people and those planning to get pregnant to get vaccinated against COVID-19. The known and generated responses both exhibit themes of concern for the effects on unborn children (shown in \textcolor{red}{red}), concern for the vaccines getting FDA approval (shown in \textcolor{brown}{brown}), and feelings of encouragement toward the recommendation (shown in \textcolor{blue}{blue}). The fourth example shows a test message from Vaccines  where CDCDirector discusses updating pediatric vaccine recommendations to include children 5-11 years old. The known and generated responses both exhibit feelings of gratefulness and acknowledgement of the importance of pediatric vaccination (shown in \textcolor{red}{red}).

\subsection{Envisioned Use Case Walk-through}\label{section_envisioned_use_case}

\begin{figure*}[t]
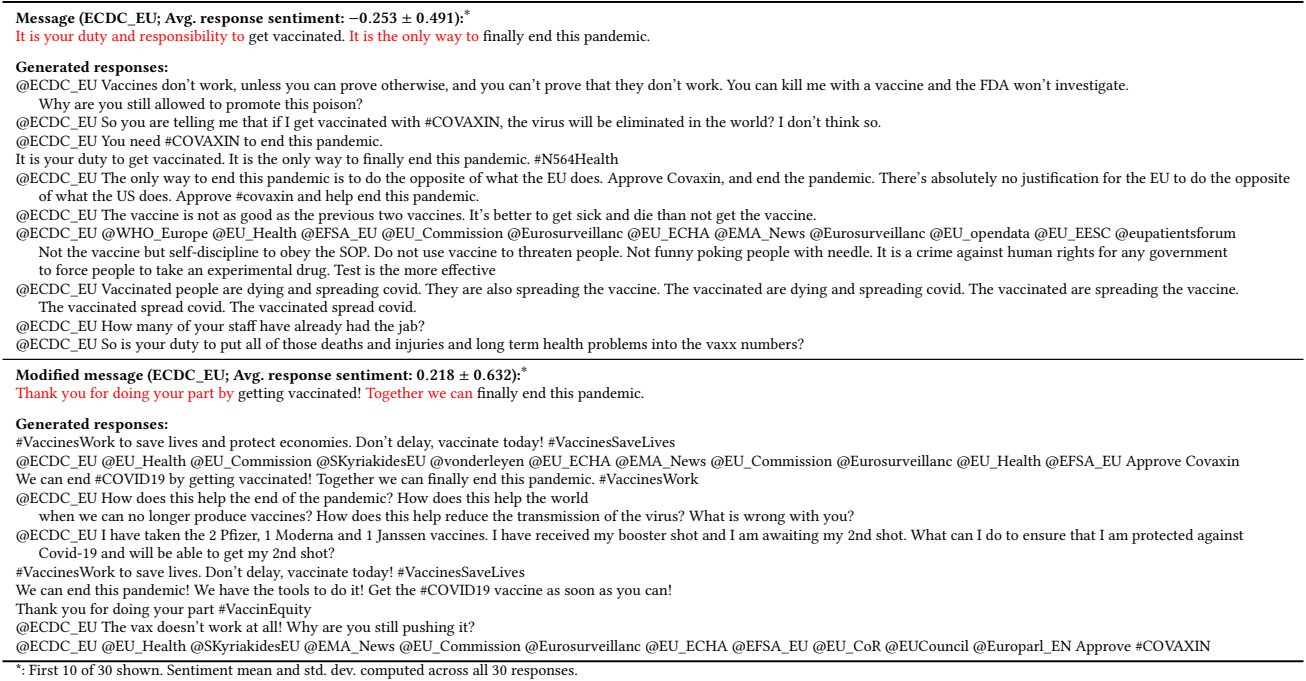

\begin{center}
\scriptsize
\begin{tabular}{l}
    \toprule
    \textbf{Message (ECDC\_EU; Avg. response sentiment: $\mathbf{-0.253\pm 0.491}$):$^*$}\\
    \textcolor{red}{It is your duty and responsibility to} get vaccinated. \textcolor{red}{It is the only way to} finally end this pandemic. \\
    \addlinespace
    \textbf{Generated responses:}\\
	@ECDC\_EU Vaccines don't work, unless you can prove otherwise, and you can't prove that they don't work. You can kill me with a vaccine and the FDA won't investigate. \\
	\hspace{0.1in} Why are you still allowed to promote this poison?\\
	@ECDC\_EU So you are telling me that if I get vaccinated with \#COVAXIN, the virus will be eliminated in the world? I don't think so.\\
	@ECDC\_EU You need \#COVAXIN to end this pandemic.\\
	It is your duty to get vaccinated. It is the only way to finally end this pandemic. \#N564Health\\
	@ECDC\_EU The only way to end this pandemic is to do the opposite of what the EU does. Approve Covaxin, and end the pandemic. There's absolutely no justification for the EU to do the opposite \\
	\hspace{0.1in} of what the US does. Approve \#covaxin and help end this pandemic.\\
    @ECDC\_EU The vaccine is not as good as the previous two vaccines. It's better to get sick and die than not get the vaccine.\\
	@ECDC\_EU @WHO\_Europe @EU\_Health @EFSA\_EU @EU\_Commission @Eurosurveillanc @EU\_ECHA @EMA\_News @Eurosurveillanc @EU\_opendata @EU\_EESC @eupatientsforum \\
	\hspace{0.1in} Not the vaccine but self-discipline to obey the SOP. Do not use vaccine to threaten people. Not funny poking people with needle. It is a crime against human rights for any government\\
	\hspace{0.1in} to force people to take an experimental drug. Test is the more effective\\
	@ECDC\_EU Vaccinated people are dying and spreading covid. They are also spreading the vaccine. The vaccinated are dying and spreading covid. The vaccinated are spreading the vaccine. \\
	\hspace{0.1in} The vaccinated spread covid. The vaccinated spread covid.\\
	@ECDC\_EU How many of your staff have already had the jab?\\
	@ECDC\_EU So is your duty to put all of those deaths and injuries and long term health problems into the vaxx numbers?\\
    \midrule
    \textbf{Modified message (ECDC\_EU; Avg. response sentiment: $\mathbf{0.218\pm 0.632}$):$^*$} \\
    \textcolor{red}{Thank you for doing your part by} getting vaccinated! \textcolor{red}{Together we can} finally end this pandemic. \\
    \addlinespace
    \textbf{Generated responses:}\\
	\#VaccinesWork to save lives and protect economies. Don't delay, vaccinate today! \#VaccinesSaveLives\\
	@ECDC\_EU @EU\_Health @EU\_Commission @SKyriakidesEU @vonderleyen @EU\_ECHA @EMA\_News @EU\_Commission @Eurosurveillanc @EU\_Health @EFSA\_EU Approve Covaxin\\
	We can end \#COVID19 by getting vaccinated! Together we can finally end this pandemic. \#VaccinesWork\\
	@ECDC\_EU How does this help the end of the pandemic? How does this help the world\\
	\hspace{0.1in} when we can no longer produce vaccines? How does this help reduce the transmission of the virus? What is wrong with you?\\
	@ECDC\_EU I have taken the 2 Pfizer, 1 Moderna and 1 Janssen vaccines. I have received my booster shot and I am awaiting my 2nd shot. What can I do to ensure that I am protected against\\
	\hspace{0.1in} Covid-19 and will be able to get my 2nd shot?\\
    \#VaccinesWork to save lives. Don't delay, vaccinate today! \#VaccinesSaveLives\\
	We can end this pandemic! We have the tools to do it! Get the \#COVID19 vaccine as soon as you can!\\
	Thank you for doing your part \#VaccinEquity\\
	@ECDC\_EU The vax doesn't work at all! Why are you still pushing it?\\
	@ECDC\_EU @EU\_Health @SKyriakidesEU @EMA\_News @EU\_Commission @Eurosurveillanc @EU\_ECHA @EFSA\_EU @EU\_CoR @EUCouncil @Europarl\_EN Approve \#COVAXIN\\
	\bottomrule
	*: First 10 of 30 shown. Sentiment mean and std. dev. computed across all 30 responses.\\
\end{tabular}
\caption{\small Use case: ECDC\_EU considers a ``future'' message for public release. The Vaccines model anticipates a more positive reception after the message is reworded to reduce directness and implication of personal responsibility. The modified message text is highlighted in \textcolor{red}{red}.}
\label{example_modifying_message}
\end{center}
\end{figure*}

We anticipate that public health organizations may find value in being able to ``preview'' public reception to any potential message on health policy or recommendations. As such, we envision our method being incorporated as a social media insights tool that can help avoid negative receptions where possible with the aim to improve adherence to health guidance. 

To demonstrate this use case, we invent a tweet encouraging vaccination against COVID-19 and we suppose it is being considered by the ECDC\_EU content manager for future public release. We first use the generator model trained on the Vaccines dataset to predict a set of 30 responses. We then modify the tone of the message to reduce directness and implication of personal responsibility and use it to generate a new set of 30 responses. We use the RoBERTa sentiment classifier to score each response in both sets and compute the mean and standard deviation over the scores in each set. In Figure \ref{example_modifying_message}, we show the effect of the modification: the mean sentiment increases by 0.47 on the scale of [-1, 1]. The standard deviation also increases, indicating that the responses continue to retain examples of negative concerns (they just become less prevalent). We highlight the modified portion of the message in \textcolor{red}{red} and show the first ten examples from each set to help illustrate the achieved difference.

The proposed methods may also be generalized beyond public health - any organization with a presence on Twitter may tailor our method to their requirements by indexing their existing tweets and their responses in Elasticsearch and then fine-tuning GPT-2. We also note that our method is easily adaptable to other social media platforms beyond Twitter, as long as a mechanism exists in the platform for public response (e.g., Reddit).

\section{Quantitative Results}\label{section_quantitative_results}

We now describe in detail our statistical testing, the purpose of which is to confirm that our models capture the true semantic and sentiment distributions of known responses as we expect.

\subsection{Semantic Similarity}\label{section_semantic_similarity}
For each test message, we aim to establish if the model generates responses that capture the semantics (e.g., meanings, topics, intents) present in the known responses. To do so, we compute the max pairwise cosine similarity between the sentence embedding of each known primary ground-truth response and those of the reference, generated, and random responses. This yields three sets of 30 max cosine similarity values for each test message - one for the ground-truth baseline, one for the model evaluation, and one for the random-chance baseline. We choose max instead of mean cosine similarity so that primary ground-truth responses will be considered ``covered'' by the model if at least one similar response shows up in the generated sample \cite{gupta-etal-2019-investigating}. We then perform three statistical tests on each set to compare the model with the baselines: (1) the Area Under Regression Error Characteristic Curve (AUC-REC) \cite{bi2003regression} to compare the expected
cosine similarity error for the model and baselines; (2) a two-tailed paired t-test to compare the average max cosine similarity between the model and baselines; and (3) a Pearson's correlation between the max cosine similarity values of the model and those of the baselines.

\subsubsection{AUC-REC}\label{section_auc_rec}

We introduce the AUC-REC approach for assessing semantic similarity of the primary, reference, generated, and random response sets. Regression Error Characteristic (REC) curves generalize the principles behind the Receiver Operator Characteristic (ROC) curves to regression models \cite{bi2003regression}. The ROC curve is typically used to present the quality of a binary classification model by comparing its true-positive rate (along the y-axis) to its false-positive rate (along the x-axis). The area under the resulting curve (AUC-ROC) is a metric that summarizes the extent to which the classifier can correctly identify positive examples without mistaking negative examples as positive. The REC curve applies a similar premise to regression models: for each of an increasing series of error tolerances (along the x-axis) it shows the ``accuracy'' of the model within that tolerance (along the y-axis). Specifically, the accuracy is the percentage of examples for which the continuous target value can be predicted within the given error tolerance. The area over the resulting curve approximates the total expected error of the model, and thus the area under the curve can be used to approximate model quality in the same manner as ROC curves. 

\begin{figure}[t]
    \vskip 0.05in
    \begin{center}
        \hspace*{-0.5cm}
        \includegraphics[width=1.07\columnwidth]{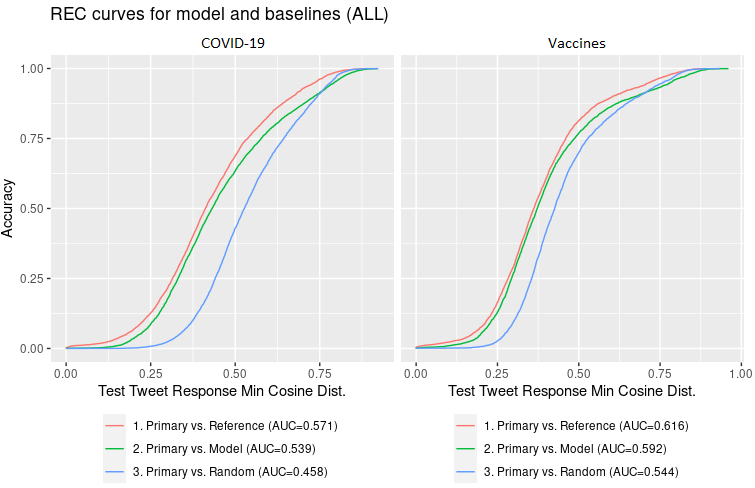}
        \caption{\small REC curves using the min cosine distance error metric on the full test sets of the COVID-19 (Left) and Vaccines (Right) datasets.}
        \label{covid-vaccines-cossim-rec}
    \end{center}
    \vskip -0.2in
\end{figure}

We use the REC curves to directly compare the ground-truth baseline (Primary vs. Reference), the model evaluation (Primary vs. Model), and the random-chance baseline (Primary vs. Random) using min cosine distance as the error metric. We construct each REC curve as follows:
(1) we concatenate the sets of 30 max cosine similarity scores for each of $M$ test messages, yielding a single list of cosine similarities for all $M \times 30$ primary ground-truth responses (e.g., for the COVID-19 dataset, this yields $155 \times 30 = 4,650$ max cosine similarities); (2) we normalize the resulting list so that the highest score is 1; and (3) we subtract all values in the list from 1 to convert them to cosine distances. All three resulting lists (one for the model evaluation and two for the baselines) are then used to construct the REC curves and AUC values as described in \cite{bi2003regression}.
Figure \ref{covid-vaccines-cossim-rec} shows the curves with corresponding AUC measurements for the model and baselines on both datasets. In Table \ref{semantic_similarity_auc_rec_results} we report the AUC scores for the full test set (ALL) and report them again separately for each twitter account with at least 20 messages in the test set of both datasets (WHO, CDCgov, CDCDirector). REC plots for these individual accounts are provided in Appendix \ref{section_additional_rec_plots}.

\begin{table}[t]
    \centering
    \footnotesize
    \caption{\small Semantic Similarity: AUC-REC Results}
    \label{semantic_similarity_auc_rec_results}
    \begin{tabular}{lllll}
        \toprule
        \multicolumn{5}{l}{\textbf{COVID-19 dataset}}\\
        \midrule
        \textbf{Comparison} & \textbf{ALL} & \textbf{WHO} & \textbf{CDCgov} & \textbf{CDCDirector}\\
        \midrule
        Primary vs. Reference   & 0.571  & 0.565  & 0.558  & 0.610 \\
        Primary vs. Model       & 0.539  & 0.517  & 0.544  & 0.595 \\
        Primary vs. Random      & 0.458  & 0.442  & 0.466  & 0.500 \\
        \midrule
        Model \% Difference$^*$ & 71.7\% & 61.0\% & 84.8\% & 86.4\% \\
        \toprule
        \multicolumn{5}{l}{\textbf{Vaccines dataset}}\\
        \midrule
        \textbf{Comparison} & \textbf{ALL} & \textbf{WHO} & \textbf{CDCgov} & \textbf{CDCDirector}\\
        \midrule
        Primary vs. Reference   & 0.616  & 0.653  & 0.599  & 0.626 \\
        Primary vs. Model       & 0.592  & 0.620  & 0.576  & 0.609 \\
        Primary vs. Random      & 0.544  & 0.546  & 0.538  & 0.559 \\
        \midrule
        Model \% Difference$^*$ & 66.7\% & 69.2\% & 62.3\% & 74.6\% \\
        \bottomrule
        \multicolumn{5}{l}{\scriptsize *: Model \% Difference: 100 * (Model - Random) / (Reference - Random)}\\
    \end{tabular}
    \vskip -0.15in
\end{table}

To compare model performance across datasets and test accounts, we compute the Model \% Difference, which is the position of the model evaluation AUC relative to the distance between the upper and lower bounds established by the two baselines (e.g., 100\% indicates model equals reference, and 0\% indicates model equals random).
Note that for both datasets and for each account, the min cosine distance AUC for the model evaluation is much closer to that of the ground-truth baseline than to that of the random-chance baseline (e.g., Model \% Difference = 71.7\% for COVID-19 and 66.7\% for Vaccines). 
This indicates that the model is able to capture and reproduce the true semantics of typical responses to messages and authors in our test sets. In the COVID-19 dataset, the model had an easier time reproducing the semantic content of responses to the CDCgov and CDCDirector accounts compared to the WHO and account (e.g., Model \% Difference = 86.4\% for CDCDirector, 84.8\% for CDCgov, and only 61.0\% for WHO). However in the Vaccines dataset, the model had the easiest time with CDCDirector, followed by WHO and then CDCgov (e.g., Model \% Difference = 74.6\% for CDCDirector, 69.2\% for WHO, and only 62.3\% for CDCgov).  

\subsubsection{Paired t-Tests}\label{section_paired_t_tests}
We follow up the REC-AUC analysis with confirmatory two-tailed paired t-tests to directly compare the differences in average max cosine similarity between the model evaluation and the baselines. We again concatenate the sets of 30 max cosine similarity scores for each of $M$ test messages, 
yet this time we do not normalize them or convert them to cosine distance. This yields three lists of $M \times 30$ max cosine similarities (one for the model evaluation and two for the baselines), and we run two t-tests: (1) comparing difference in mean between the lists for both baselines, and (2) comparing the difference in mean between the model evaluation list and the random-chance baseline list. Each test is run with the null hypothesis that there is no difference between the means of the lists, giving a p-value at the 5\% significance level for any observed differences. 

In Table \ref{semantic_similarity_t_test_results} we report the results of these tests for both datasets. We again report results for each full test set (ALL) and breakdowns for each twitter account with at least 20 messages in the test sets (WHO, CDCgov, CDCDirector). 
Also, as done previously for AUC-REC, we compare model performance across datasets and test accounts using Model \% Difference. This time we do so using the differences in means for max cosine similarity confirmed via the t-tests. We observe an absolute difference of less than 1\% between the Model \% Differences obtained for the paired t-tests and those obtained for the AUC-REC scores (e.g., on the full COVID-19 test set we have Model \% Difference = 71.7\% for AUC-REC and 70.8\% for the paired t-tests, and on the full Vaccines test set we have Model \% Difference = 66.7\% for AUC-REC and 67.6\% for the paired t-tests). This provides confirmation for the conclusions drawn from the AUC-REC results; that is, that the model can meaningfully capture and reproduce response semantics for test messages and authors. 

\begin{table}[t]
    \centering
    \footnotesize
    \caption{\small Diff. in Mean for max cosine Sim. (paired t-test)}
    \label{semantic_similarity_t_test_results}
    \begin{tabular}{lllll}
        \toprule
        \multicolumn{5}{l}{\textbf{COVID-19 dataset} (all results significant at $p \ll 0.01$)}\\
        \midrule
        \textbf{Comparison} & \textbf{ALL} & \textbf{WHO} & \textbf{CDCgov} & \textbf{CDCDirector}\\
        \midrule
        GT vs. Random Baselines   & +0.113  & +0.124  & +0.093  & +0.110 \\
        ME vs. Random Baseline    & +0.080  & +0.076  & +0.078  & +0.095 \\
        \midrule
        Model \% Difference$^*$   & 70.8\%  & 61.3\%  & 83.9\%  & 86.4\% \\
        \toprule
        \multicolumn{5}{l}{\textbf{Vaccines dataset} (all results significant at $p \ll 0.01$)}\\
        \midrule
        \textbf{Comparison} & \textbf{ALL} & \textbf{WHO} & \textbf{CDCgov} & \textbf{CDCDirector}\\
        \midrule
        GT vs. Random Baselines   & +0.071  & +0.108  & +0.061  & +0.067 \\
        ME vs. Random Baseline    & +0.048  & +0.074  & +0.038  & +0.050 \\
        \midrule
        Model \% Difference$^*$   & 67.6\%  & 68.5\%  & 62.3\%  & 74.6\% \\
        \bottomrule
        \multicolumn{5}{l}{\scriptsize *: Model \% Difference: 100 * (ME vs. Random Baseline) / (GT vs. Random Baselines)}\\
        \multicolumn{5}{l}{\scriptsize GT = Ground-truth; ME = Model Evaluation}\\
    \end{tabular}
    \vskip -0.05in
\end{table}

\subsubsection{Correlation}\label{section_correlation}

\begin{figure}[t]
    \begin{center}
        \includegraphics[width=\columnwidth]{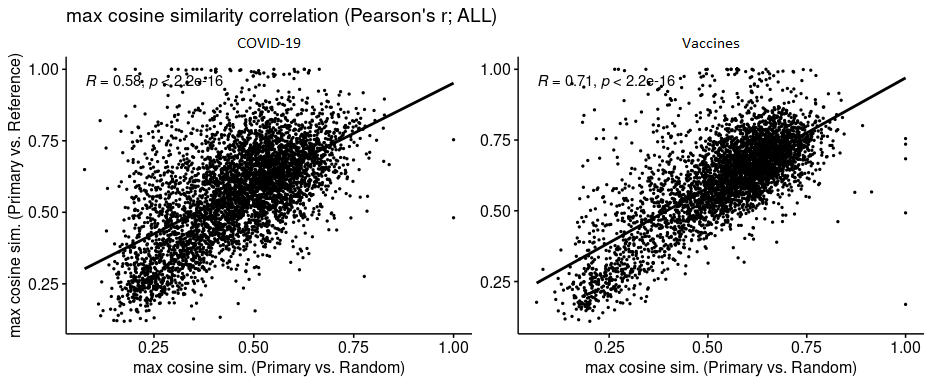}
        \vskip -0.05in
        \caption{\small Pearson correlation between max cosine similarities of the ground-truth baseline and the random-chance baseline on the full test sets of the COVID-19 (Left) and Vaccines (Right) datasets. }
        \label{cossim_corr}
    \end{center}
    \vskip -0.1in
\end{figure}

Finally, we perform a correlation study between the max cosine similarity scores of the ground-truth baseline (Primary vs. Reference) and those of the random-chance baseline (Primary vs. Random). 
The purpose of this study is to identify the base level of semantic relatedness that any pair of random responses (to any message) has in each dataset, and investigate the degree to which this increases for pairs of responses to the same messages. This captures the difficulty inherent in learning to predict semantics conditional on individual messages and authors. For example, imagine a degenerate dataset in which all responses are the same regardless of the message;  in such a scenario there would not be much for the model to learn, and we would see a perfect linear correlation between the two baselines.

We use the same concatenated lists of $M \times 30$ max cosine similarities used in the t-tests, this time only using the ones for the ground-truth and random-chance baselines. For each dataset, we compute the Pearson's correlation coefficient $r$ between these two lists. 
As seen in Figure \ref{cossim_corr}, we observe that  COVID-19  has more semantically diverse responses with correlation $r=0.58$ p-value $< 2.2\times 10^{-16}$  between the ground-truth and random-chance baselines, while Vaccines is much less so with $r=0.71$ p-value $< 2.2\times 10^{-16}$  between baselines. This indicates that Vaccines presents an ``easier'' problem for the model with respect to learning semantic distributions. This explains why model evaluation metrics are better for Vaccines (e.g., lower validation perplexity, higher AUC) than for the COVID-19 dataset, yet we see higher Model \% Differences for COVID-19. Although we have already established using the AUC-REC and t-test analysis that GPT-2 is effective at generating semantically correct response distributions on both datasets, this correlation analysis shows that use of such a model has more \textit{utility} on the COVID-19 dataset than on the Vaccine dataset. When considering how a newly authored COVID-19 related tweet would be received, a user is less likely to find accurate insight by simply looking at related historical responses and would benefit more from a generative model. 
 
\subsection{Sentiment Similarity}\label{section_sentiment_similarity}

\begin{table}[t]
    \centering
    \footnotesize
    \caption{\small Sentiment Similarity: Results of pair wise Chi-Square tests}
    \label{sentiment_similarity_chi_sq_results}
    \begin{tabular}{lllll}
        \toprule
        \multicolumn{5}{l}{\textbf{COVID-19 dataset}}\\
        \midrule
        \textbf{Comparison} & \textbf{ALL} & \textbf{WHO} & \textbf{CDCgov} & \textbf{CDCDirector}\\
        \midrule
        Primary vs. Reference   & 72.9\%  & 77.1\%   & 70.4\%   & 65.4\%  \\
        Primary vs. Model       & 55.5\%   & 49.4\%   & 65.9\%   & 57.7\%  \\
        Primary vs. Random      & 43.9\%   & 41.0\%   & 65.9\%   & 19.2\%  \\
        \toprule
        \multicolumn{5}{l}{\textbf{Vaccines dataset}}\\
        \midrule
        \textbf{Comparison} & \textbf{ALL} & \textbf{WHO} & \textbf{CDCgov} & \textbf{CDCDirector}\\
        \midrule
        Primary vs. Reference   & 63.6\%  & 82.6\%   & 59.1\%   & 58.1\%  \\
        Primary vs. Model       & 52.8\%   & 34.8\%   & 57.7\%   & 55.8\%  \\
        Primary vs. Random      & 43.6\%   & 43.5\%   & 45.0\%   & 44.2\%  \\
        \bottomrule
    \end{tabular}
    \vskip -0.1in
\end{table}

Having established that the model effectively generates semantically similar responses to messages from the different accounts, we now analyze the sentiments that are reflected by the modeled responses and compare them against the sentiments reflected in Primary, Reference and Random responses. We assess if the sentiments expressed by the Model and the Primary, Reference and Random populations are distributed similarly. 

As discussed in Section \ref{section_experimental_setup}, we use RoBERTa to assign sentiment scores for each response. We bin the score ($s$) of each primary, reference, generated, and random response into three classes: (1) Negative, where $1 \leq s < -0.25$, (2) Neutral, where $-0.25 \leq s \leq 0.25$, and (3) Positive, where $0.25 < s \leq 1$.
We then perform three Chi-square tests for each test message to compare the class distribution of its primary ground-truth responses and those of its reference, generated, and random responses. The Chi-squared statistic  represents the difference that exists between observed and expected counts, if there is no relationship in the population. The null hypothesis of each test assumes there is no difference in class distribution, and the p-value gives the probability that any observed differences are due to chance. This yields three p-values for each message - one for the ground-truth baseline, one for the model evaluation, and one for the random-chance baseline. The percentage of messages where we fail to reject the null hypothesis with a significance level of 5\% is counted for the model and the baselines. These percentages reflect the proportion of messages for which there is no significant difference in the sentiment distribution between the compared sets.

In Table \ref{sentiment_similarity_chi_sq_results} we report the percentage of test tweets for which there is no significant difference in sentiment distribution on the basis of failures to reject the null hypothesis in pair wise Chi-Square tests, for Primary vs. Reference, Primary vs. Model and Primary vs Random comparison sets, for both the COVID19 and Vaccine datasets. Analyzing the percentage values in each column for the organizations, the Primary vs. Reference comparison set provides the greatest match in sentiments distributions (72.9\% for COVID-19 and 63.6\% for Vaccines, for ALL organizations) followed by Primary vs. Model (55.5\% for COVID-19 and 52.8\% for Vaccines, for ALL organizations). The Model has a greater match of sentiments than the Primary vs. Random test (43.9\% for COVID-19 and 43.6\% for Vaccines, for ALL organizations). 

Thus, the sentiment analysis results on the model-generated responses reflect that the model mostly captures the sentiment distributions of the known ground-truth responses. Only in one instance, Vaccine data set for WHO, the model generated responses yield a worse percentage than Random when compared against the Primary sentiment distribution. 

\begin{figure}[t]
    \begin{center}
        \includegraphics[width=0.95\columnwidth]{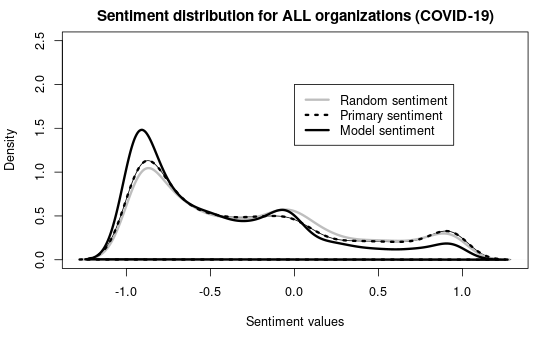}
        \includegraphics[width=0.95\columnwidth]{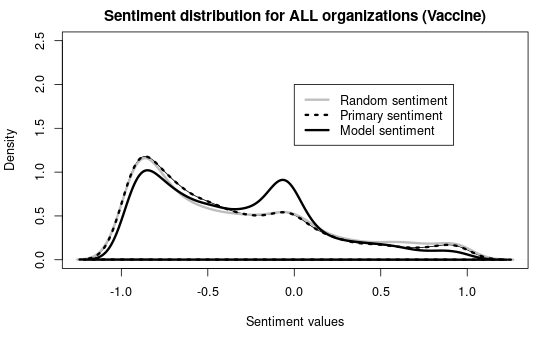}
        \vskip -0.1in
        \caption{\small Distribution of sentiments including all organizations, for COVID 19 data (Top) and Vaccine  data (Bottom)}
        \label{All Sentiments}
    \end{center}
    \vskip -0.1in
\end{figure}

To further investigate how close the sentiment values from the Model, Primary and Random responses are, we looked at the density distribution of the raw sentiment values from RoBERTa for ALL organizations.
Figure \ref{All Sentiments} represents the density distribution of the sentiment scores provided for the Primary, Model (generated) and Random responses for ALL tweets for each data set. 

\begin{figure}[t] 
    \begin{center}
        \includegraphics[width=0.95\columnwidth]{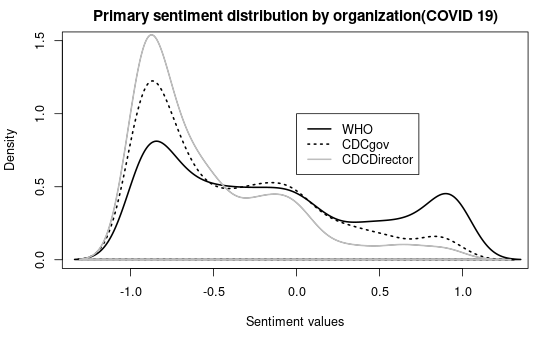}
        \includegraphics[width=0.95\columnwidth]{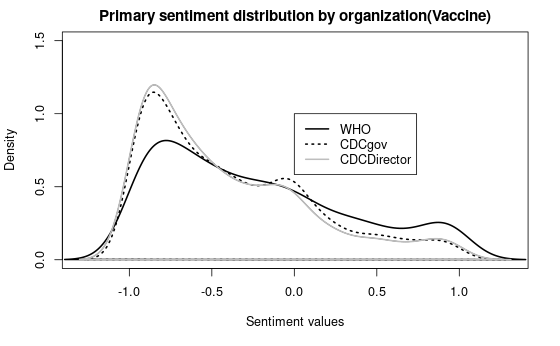}
        \vskip -0.1in
        \caption{\small Distribution of sentiments for individual organizations, for COVID 19 data (Top) and Vaccine data (Bottom)}
        \label{Organization Sentiments}
    \end{center}
    \vskip -0.1in
\end{figure}

The density distribution of sentiments from the Primary, Model and Random responses reflect highest density peaks for negative sentiments (peaking close to sentiment value of -1.0). To understand if this is due to the relative differences of public message reception from different organizations, we investigate the density distribution obtained from the sentiments from Primary ground truth messages and responses for each  public health organization  in Figure \ref{Organization Sentiments}.

We note that there seem to be more negative sentiments in the  ground truth responses for CDCgov and CDCDirector accounts, when compared with that for the WHO. It is important to note that our models are text (response) generators and not directly trained to predict sentiment class likelihood. Also, since the models are not trained separately for each organization, the relative differences in response sentiments between WHO and other organizations may contribute to the diminished performance we observe capturing the true sentiment distribution in responses to WHO messages (as reflected in results from Vaccine data in Table \ref{sentiment_similarity_chi_sq_results}).

\section{Related
Work}\label{section_related_work}
We review relevant works which introduce methods for generating social media text (e.g., tweets), or which use social media text as a basis for learning to generate conversational responses. DialoGPT \cite{zhang-etal-2020-dialogpt} is a GPT-2-based dialogue response generator trained on 147 million ``conversations'' constructed as distinct paths through comment chains on Reddit. PLATO \cite{bao-etal-2020-plato}, PLATO-2 \cite{bao-etal-2021-plato}, and BlenderBot \cite{roller-etal-2021-recipes} are recent open-domain neural conversational models that also use social media responses in pre-training (PLATO uses both Reddit and Twitter, the others use Reddit only). Cetinkaya et al. \cite{ccetinkaya2020developing} propose a neural encoder-decoder-reranker framework for building chatbots that can participate in Twitter conversations by learning to take a side on a controversial issue (e.g., gun control). Tran \& Popowich and Roy et al. both explore methods for generating tweets to notify the public about traffic incidents \citep{tran-popowich-2016-automatic, 10.1007/978-981-16-4807-6_24}. Lofi \& Krestel propose a method to use open government data to generate tweets that inform the public on ongoing political processes \cite{10.1007/978-3-642-29035-0_24}. Finally, in perhaps the most related work to ours, Garvey et al. \cite{GARVEY2021113497} propose a system designed to aid social media content managers design tweets that will be received well and become popular. Their system includes semantic and sentiment analysis components capable of estimating a tweet's target engagement, which is used in turn with a custom probabilistic generative model to synthesize tweets. Although we share the same motivations and envisioned use cases, what differentiates our work is that Garvey et al. use generative modeling to help a user craft a proposed message and assign it an estimated engagement score, while our method generates responses to a proposed message. This provides users with a view of what people might \textit{actually say} if the message goes public, offering crucial insights into the specific concerns that lead to a message being received well (or not). We believe that our methods complement Garvey et al. well - specifically, an organization which adopts both tools might craft promising candidate tweets via Garvey et al. and then preview their reception with our models.

\section{Conclusion}\label{section_conclusion}
We conclude with a summary of our contributions and a discussion of limitations, future directions, and ethical considerations.
\subsection{Contributions}	
Our main contributions are as follows: (1) we collected two datasets of public health messages and their responses on Twitter, one in context of COVID-19 and one in context of Vaccines; (2) we trained two GPT-2 text generators - one for each dataset - both capable of capturing and reproducing the semantic and sentiment distributions in known responses to public health messages; (3) we demonstrate our envisioned use case in which a public health organization uses our models to optimize expected reception for important health guidance; and (4) we introduce a novel evaluation scheme with extensive statistical testing to confirm that our models capture semantics and sentiment as we qualitatively observe.

\subsection{Limitations and Future Directions}\label{section_limitations_future_directions}
Here we note several key limitations of our study and discuss ways in which future work may address them. Specifically, we discuss the issues of: (1) factuality of generated responses; (2) quality of semantic and sentiment similarity measurement; (3) opportunities for further evaluation; and (4) generalization beyond this study.

\subsubsection{Generated Response Factuality}\label{section_generated_response_factuality}
Language models such as GPT-2 are prone to generate factually inaccurate output, often times ``hallucinating'' details (e.g., names, places, quantities, etc.) in the absence of external knowledge \cite{liu2021token}. For example, many of the generated responses in Figures \ref{example_generated_responses} and \ref{example_modifying_message} tag users and/or display hashtags that do not make sense considering the response text. Additionally, our response generator models are prone to temporal drift unless continually re-trained on up-to-date samples from Twitter. For example, our COVID-19 dataset was collected during the spring and summer of 2020 (the early months of the pandemic) and thus would not generate accurate responses to tweets concerning late-pandemic issues such as vaccine boosters, relaxed mask recommendations, and return-to-office policies.

A potential remedy for language model hallucination and temporal drift is to take advantage of recent generative models capable of integrated information retrieval from knowledge bases (e.g., RAG \cite{lewis2020retrieval}). Retrieval-augmented response generation would allow response predictions to incorporate rapidly evolving information (e.g., breaking news updates) without needing constant re-training, and could increase the general correctness of generated responses with respect to common world knowledge. Additionally, maintaining an up-to-date knowledge base of current events requires less computational resources than continually training language models.

\subsubsection{Semantic \& Sentiment Measurement Quality}\label{section_semantic_sentiment_measurement_quality}
In our study we use off-the-shelf pre-trained models for computing sentence embeddings and sentiment scores. Specifically, the MiniLM sentence embedding model was pre-trained on over one billion sentence pairs from 32 distinct datasets. These include Wikipedia, various Q\&A collections, comments from social media forums such as Reddit, Quora, Stack Exchange, and Yahoo Answers, and many others. The RoBERTa sentiment classifier was fine-tuned on the dataset used in the TweetEval sentiment analysis task, which is the SemEval 2017 Twitter sentiment analysis dataset \cite{rosenthal-etal-2017-semeval} with over 50,000 labeled English tweets. Since these models have already been exposed to large-scale corpora and perform well on their respective benchmarks, we deemed them sufficient for measuring semantic similarity and sentiment on our datasets. However, this comes with the limitation that these models have not seen specialized terminology (e.g., COVID-19) or entities (e.g., names, places) of significance beyond the topical and temporal scope of their training sets. 

To mitigate this, it is possible to do additional fine-tuning of the sentence embedding and sentiment models on the collected tweets to ensure the most robust semantic and sentiment comparisons. We encourage any future work employing our methods to explore this avenue. We note that sentence pairs for further training of the embedding model can be mined from raw tweet collections automatically (e.g., select positive example pairs from the same reply threads and negative example pairs at random), but labeling new tweets for sentiment polarity requires manual effort.

\subsubsection{Opportunities for Further Evaluation}\label{section_opportunities_for_further_evaluation}
In this work we evaluate one type of model for response generation (GPT-2). We recognize that response generation is a well studied area, specifically in conversational contexts (e.g., see Section \ref{section_related_work}), and thus there is opportunity to compare different response generation models on this task. For example, more recent, larger-scale generative models (e.g., GPT-3 \cite{brown2020language}) are likely to produce higher quality responses at the cost of increased compute for training and evaluation. However, we note that new language models are constantly being developed and improved, and our proposed methodology supports the replacement of GPT-2 with any current or future text generation model without changing the nature of the task, evaluation, or its use-cases.

We also note the need for end-user validation by our target audience (e.g., public health social media managers). This could be in the form of a trial where users use our models, perhaps in combination with the tools of Garvey et al., to respond to their messages and then compare the system's predictions with actual responses on Twitter. Such a study may yield valuable information regarding the effectiveness of our methods as day-to-day tools and produce directions for future improvement.

\subsubsection{Generalization Beyond This Study}\label{section_generalization_beyond_this_study}
Our statistical evaluations demonstrate the effectiveness of generative response modeling in reproducing the sentiment and semantics of public health responses on Twitter. However, as briefly noted in Section \ref{section_envisioned_use_case}, there is ample opportunity to generalize our method to other settings. Some possibilities for future work include: (1) allowing the response generator to be conditioned on attributes of the responder (e.g., geographical region, age, etc.) to provide insights into how targeted populations might react to a message; (2) training expanded models on additional author types beyond public health organizations (e.g., political organizations and large corporations); and (3) targeting other social media platforms (e.g., Facebook and Reddit).

\subsection{Ethical Considerations}\label{section_ethical_considerations}
We recognize the potential dangers presented by the use of language models such as GPT-2 to emulate unfiltered public discourse as we do in this study. The examples in Figure \ref{example_generated_responses} make evident the degree to which such models can be prompted to emit vitriol in this setting, and there is obvious direction for misuse. We take this opportunity to reiterate that our intended use case is to allow social media representatives for impactful organizations to gain accurate perspectives on the way their messages may be received by the public, which requires preserving the real semantics and sentiment of social media discourse regardless of its toxicity. We do not support or condone the use of our methods, models, or data for any purpose that may directly or indirectly cause harm to others.

\begin{acks}
  This study was supported by the Rensselaer Institute for Data Exploration and Applications (IDEA), the Rensselaer Data INCITE Lab, and a grant from the United Health Foundation. Additionally, we thank Brandyn Sigouin, Thomas Shweh, and Haotian Zhang for their participation in the exploratory phase of this project via the Data INCITE Lab.
\end{acks}

\bibliographystyle{ACM-Reference-Format}
\bibliography{bibliography}


\newpage
\onecolumn

\appendix

\section{Additional REC Plots}\label{section_additional_rec_plots}
\begin{figure}[h]
    \begin{center}
        \hspace*{-0.5cm}
        \includegraphics[width=\textwidth]{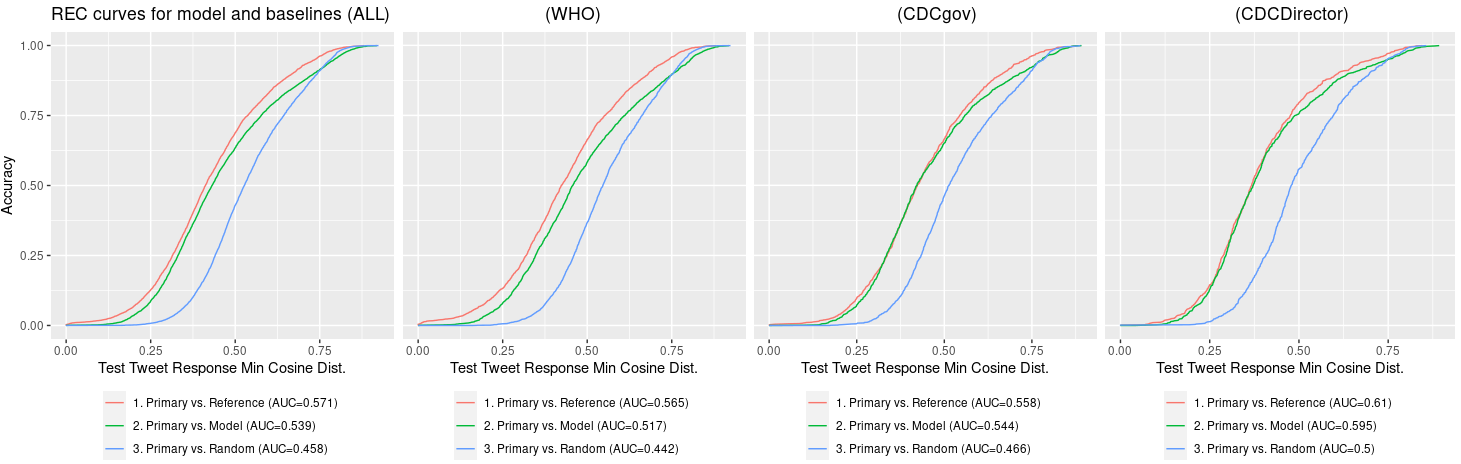}
        \caption{REC curves using the min cosine distance error metric on the test set of the COVID-19 dataset.}
        \label{covid-ph-cossim-rec}
    \end{center}
\end{figure}

\begin{figure}[h]
    \begin{center}
        \hspace*{-0.5cm}
        \includegraphics[width=\textwidth]{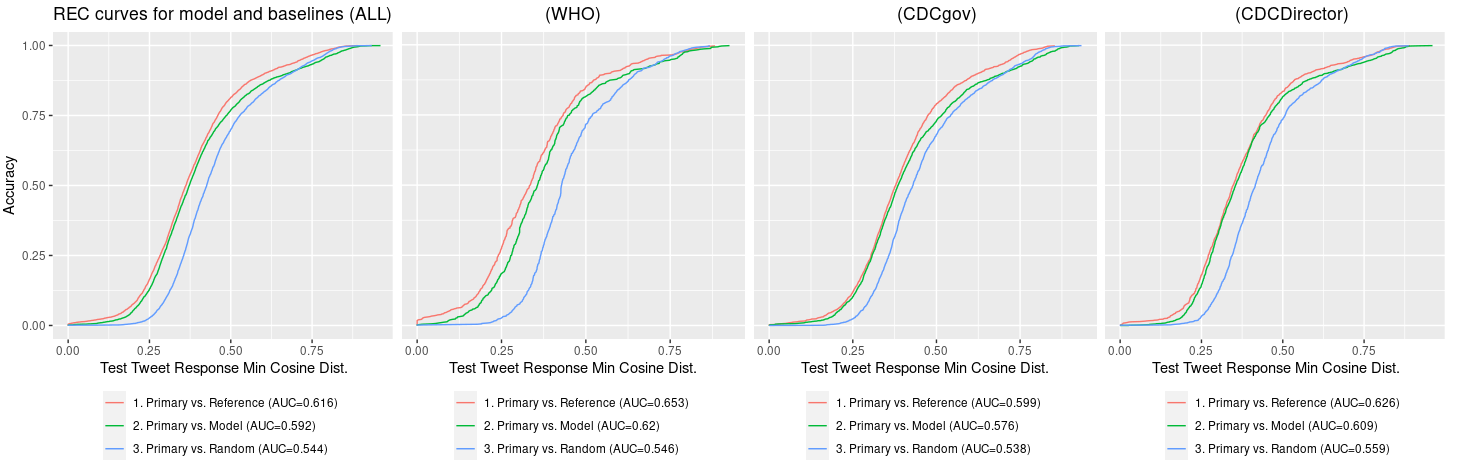}
        \caption{REC curves using the min cosine distance error metric on the test set of the Vaccines dataset.}
        \label{vaccines-ph-cossim-rec}
    \end{center}
\end{figure}

The following shows the semantic similarity REC curves for each individual public health organization account with at least 20 messages in the test set of both datasets. Figure \ref{covid-ph-cossim-rec} shows REC curves for the COVID-19 dataset and Figure \ref{vaccines-ph-cossim-rec} for the Vaccines dataset. The REC curves for each full test set (left-most plot in each figure) are provided here again to facilitate comparison.

\end{document}